\documentclass[letterpaper, 10 pt, conference]{ieeeconf}  

\IEEEoverridecommandlockouts                              

\usepackage{amsmath} 
\usepackage{amssymb}  
\usepackage{siunitx}
\usepackage{esvect}
\usepackage{multirow}
\usepackage{graphicx}
\usepackage{mathtools}
\usepackage{float}
\usepackage[font=small,skip=1pt,belowskip=1pt]{subcaption}
\usepackage[font=small,skip=1pt,belowskip=1pt]{caption}
\usepackage{cite}

\makeatletter
\let\NAT@parse\undefined
\makeatother
\usepackage[final]{hyperref} 
\hypersetup{colorlinks = true,
            linkcolor = blue,
            urlcolor  = blue,
            citecolor = blue,
            anchorcolor = blue}
\def\BibTeX{{\rm B\kern-.05em{\sc i\kern-.025em b}\kern-.08em
    T\kern-.1667em\lower.7ex\hbox{E}\kern-.125emX}}

\title{\LARGE \bf
Following Social Groups: Socially Compliant Autonomous \\ Navigation in Dense Crowds
}

\author{Xinjie Yao$^{1}$, Ji Zhang$^{2}$ and Jean Oh$^{2}$
\thanks{$^{1}$X. Yao is with the Department of Electronic and Computer Engineering, Hong Kong University of Science and Technology.
Email: {\tt xyaoab@ust.hk}}%
\thanks{$^{2}$J. Zhang and J. Oh are with the Robotics Institute, Carnegie Mellon University. Emails: {\tt zhangji@cmu.edu, jeanoh@cmu.edu}}%
}

\begin{document}

\maketitle
\thispagestyle{empty}
\pagestyle{empty}

\begin{abstract}
In densely populated environments, socially compliant navigation is critical for autonomous robots as driving close to people is unavoidable. This manner of social navigation is challenging given the constraints of human comfort and social rules. Traditional methods based on hand-craft cost functions to achieve this task have difficulties to operate in the complex real world. Other learning-based approaches fail to address the naturalness aspect from the perspective of collective formation behaviors. We present an autonomous navigation system capable of operating in dense crowds and utilizing information of social groups. The underlying system incorporates a deep neural network to track social groups and join the flow of a social group in facilitating the navigation. A collision avoidance layer in the system further ensures navigation safety. In experiments, our method generates socially compliant behaviors as state-of-the-art methods. More importantly, the system is capable of navigating safely in a densely populated area ($10+$ people in a $\SI{10}{\meter}\times\SI{20}{\meter}$ area) following crowd flows to reach the goal.
\end{abstract}

\section{Introduction}

The ability to safely navigate in populated scenes, e.g. airports, shopping malls, and social events, is essential for autonomous robots. The difficulty comes from the fact that people walk closely to the robot cutting ways in front of the robot or between the robot and the goal point. The safety margin for the robot to drive in crowded scenes is pushed to the minimum. In such a case, the navigation system has to trade-off between driving safely close to people and reaching the goal quickly. Furthermore, a previous study of socially compliant navigation \cite{thi2013survey} states three aspects in terms of the robot behaviors -- \textit{comfort} as the absence of annoyance and stress for humans in interaction with robots, \textit{naturalness} as the similarity between the robot and human behaviors, and \textit{sociability} as to abide by general cultural conventions. Among these three aspects, the first aspect essentially reflects safety of the navigation.

Previous studies on socially compliant navigation attempt to solve the problem with various methods, including data-driven approaches for human trajectory prediction \cite{alahi2017slstm,socialgan}, potential field-based \cite{hoeller2007potential} and social force model-based \cite{socialforce} approaches. In particular, reinforcement learning-based methods use reward functions to penalizes improper robot behaviors eliminating the cause of discomfort \cite{chen@2018drl, changan2019crowdrobot}. Inverse reinforcement learning based-methods learn from expert demonstrations \cite{franceschi2019irl}. These methods are hard to generalize due to that a large set of comprehensive expert demonstrations are hard to acquire.

The study of this paper is based on our previous work which uses deep learning in solving the socially compliant navigation problem \cite{navigan}. This paper extends the work in two ways. First, we consider the findings from a previous study \cite{moussaid2010group} that 70\% of people walk in social groups. Crowd behavior can be summarized as flows of social groups, and humans tend to move along the flow. It is our understanding that the behavior of joining the flow that shares similar heading direction is more socially compliant, causing fewer collisions and disturbances to surrounding pedestrians. Our method recognizes social groups and selects the flow to follow. Second, we ensure safety with a multi-layer navigation system. In this system, a deep learning-based global planning layer makes high-level socially compliant behavioral decisions while a geometry-based local planning layer handles collision avoidance at a low-level.


The paper is further related to previous work on modeling aggregate interactions among social groups \cite{moussaid2010group} and leveraging learned social relations in tracking group formations \cite{linder2014groupsvm}. Our main contributions are a deep learning-based method for socially compliant navigation with an emphasis on tracking and joining the crowd flow and an overall system integrated with the deep leaning method capable of safe autonomous navigation in dense crowds.

\section{Method}

\subsection{System Overview}

\begin{figure}[b!]
  \centering
  \vspace{-.2cm}
  \includegraphics[width=\linewidth]{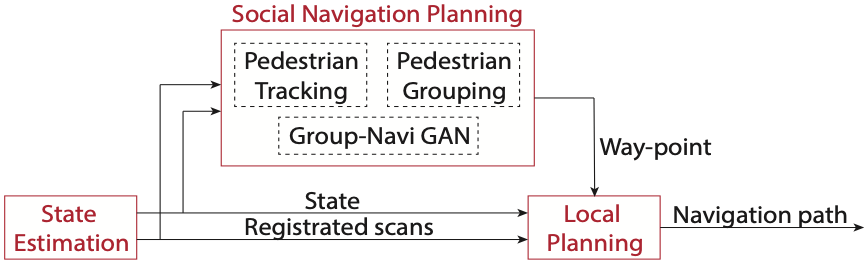}
  \vspace{.2cm}
  \caption{Navigation software system diagram.}
\label{fig:system}
\end{figure}

Fig.~\ref{fig:system} gives an overview of the autonomous navigation system which consists of three subsystems as follows.
\begin{itemize}
    \item \textbf{State Estimation Subsystem} involves a multi-layer data processing pipeline which leverages lidar, vision, and inertial sensing \cite{loam}. The subsystem computes the 6-DOF pose of the vehicle as well as registers laser scan data with the computed pose.
    \item   \textbf{Local Planning Subsystem} is a low-level planning subsystem in charge of obstacle avoidance in the vicinity of the vehicle. The planning algorithm involves a trajectory library and computes collision-free paths for the vehicle to navigate \cite{localplanner}.
    \item  \textbf{Social Navigation Planning Subsystem} takes in observations only consisting of pedestrians by subtracting the prior map. The subsystem tracks pedestrians in the surroundings of the vehicle, and then extracts the grouping information from the pedestrian walking patterns, with which, the subsystem generates way-points (as input of the Local Planning Subsystem), leveraging Group-Navi GAN, a generative planning algorithm in an adversarial training framework based on a deep neural network, Navi-GAN \cite{navigan}. 
\end{itemize}

\subsection{Group-Navi GAN}

Following the extended social force model \cite{moussaid2010group}, we propose Group-Navi GAN, a framework to jointly address the safety and naturalness aspects at a group's level. Group-Navi GAN is inspired by our previous work Navi-GAN \cite{navigan} which models social forces at an individual's level. An intention-force generator in the Group-Navi GAN deep network models the driving force as $\vv{f_i}^{\,0}$ for target agent $i$ to move toward the goal. A group-force generator models the repulsive force from other pedestrians j as $\vv{f_{ij}}^{\, }$ and the interaction force from other group members as $\vv{f_{i}}^{\, group}$. The joint output of the intention-force generator and group-force generator defines the path for the robot to navigate.

\begin{figure}[h]
  \centering
  \includegraphics[width=0.8\linewidth]{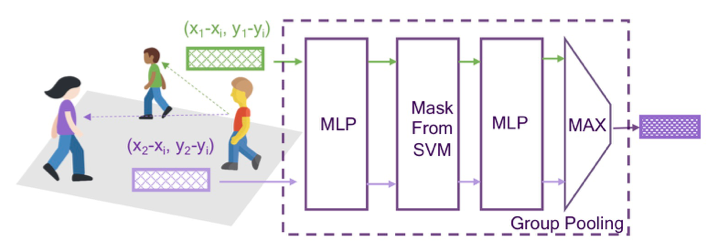}
  \vspace{.2cm}
  \caption{Group pooling module in the Group-Navi GAN deep network. The input of the module is the relative displacements of the surrounding pedestrians w.r.t. the target agent. The module associates the target agent to a group based on the motion information and outputs path adjustments for the robot to follow the group.}
\label{fig:pooling}
  \vspace{-.5cm}
\end{figure}

In the group-force generator, a group pooling module first associates the target agent to a group based on the motion information (see Fig.~\ref{fig:pooling}). Then, the group pooling module computes path adjustments which essentially guide the robot to follow the group. We apply a support vector machine classifier \cite{cortes2016svnetworks} trained by \cite{linder2014groupsvm} to determine if two agents belong to the same group. This uses the local spatio-temporal relation to cluster the agents with similar motions based on the coherent motion indicators, i.e. the differences in walking speed, spatial locations, and headings.

We use the following equation to aggregate the hidden state from ${h_j}^t$ to ${h'_j}^t$,
\begin{equation}
{h'_j}^t = I_{ij}[s_i = s_j] * cos(\theta_i - \theta_j) * {h_j}^t,
\end{equation}
where $I_{ij}[s_i = s_j]$ indicates if two agents are in the same group,
\begin{align}
    \begin{aligned}
    I_{ij}[s_i = s_j] &=
    \begin{cases}
      1, & \text{if $i$ and $j$ are in the same group} \\
      0, & \text{otherwise}
    \end{cases} \\
    \end{aligned}
\end{align}
$\theta_i$ and $\theta_j$ are the agent headings. The resulting embedding $H_i^t$ of hidden state ${h'_j}^t$ is computed as a row vector which consists of the maximum elements from all other agents. The embedding is further concatenated for decoding,
\begin{equation}
{H'_i}^t = [{H_i}^t, {h_i}^t, n_i]
\end{equation}
where $n_i$ is random noise drawn from $\mathcal{N}(0,1)\,$.

\begin{table}[b!]
\begin{center}
\resizebox{\linewidth}{!}{
\begin{tabular}{c|c|c||c|c|c|c}
Metric & Dataset & Group Percentage & Linear & SGAN \cite{socialgan}& Navi-GAN\cite{navigan} & Group-Navi GAN \\
\hline \hline
\multirow{6}{*}{ ADE } & \textbf{ETH}\cite{eth} & 18\% & 0.84  & \textbf{0.60}  & 0.95 & 1.33 \\
 & \textbf{HOTEL}\cite{eth} & 19\% & \textbf{0.35}  & 0.48 & 0.43 & 0.39   \\
 & \textbf{UNIV}\cite{ucy}& \textbf{73\%} & 0.56  & 0.36  & 0.85 & 
\textbf{0.29}  \\
 & \textbf{ZARA1}\cite{ucy} & \textbf{70\%} & 0.41 & 0.21  & 0.40  & \textbf{0.21} \\
 & \textbf{ZARA2}\cite{ucy} & 69\% & 0.53 & \textbf{0.27}  & 0.47 & 0.30 \\
\hline
 & \textbf{AVG} & 50\%  &  0.54 & 0.39 & 0.62 & 0.50 \\
\hline 
\hline
\multirow{6}{*}{ FDE } & \textbf{ETH}\cite{eth} & 18\% & 1.60  & \textbf{1.22}  & 1.64 & 1.98 \\
 & \textbf{HOTEL}\cite{eth}& 19\% & \textbf{0.60}  & 0.95 & 0.74 & 0.93 \\
 & \textbf{UNIV}\cite{ucy}& \textbf{73\%} & 1.01  & 0.75  & 1.36 &
\textbf{0.68} \\
 & \textbf{ZARA1}\cite{ucy}& \textbf{70\%} & 0.74 & 0.42 & 0.66 & \textbf{0.40} \\
 & \textbf{ZARA2}\cite{ucy}& 69\% & 0.95 & \textbf{0.54} & 0.72 & 0.85 \\
\hline
& \textbf{AVG} & 50\% & 0.98 & 0.78 & 1.02 &  0.96 \\
\end{tabular}}
\end{center}
\vspace{.2cm}
\caption{Social compliance evaluation of Group-Navi GAN and other baseline approaches. Two error metrics, Average Displacement Error and Final Displacement Error are reported (in meters) for $t_{obs}=8$ and $t_{pred}=8$. We manually count the number of pedestrians moving in social groups. Our method outperforms the prior work with the UNIV and ZARA1 datasets where social groups are richly available.}
\label{Table:metrics}
\end{table}

\setcounter{figure}{2}

\begin{figure*}[h]
    \centering
    \rotatebox[origin=c]{90}{\parbox{2.8cm}{{\footnotesize Without Social Model}}}\;\raisebox{-0.5\height}{\includegraphics[height=3cm,width=.2\linewidth]{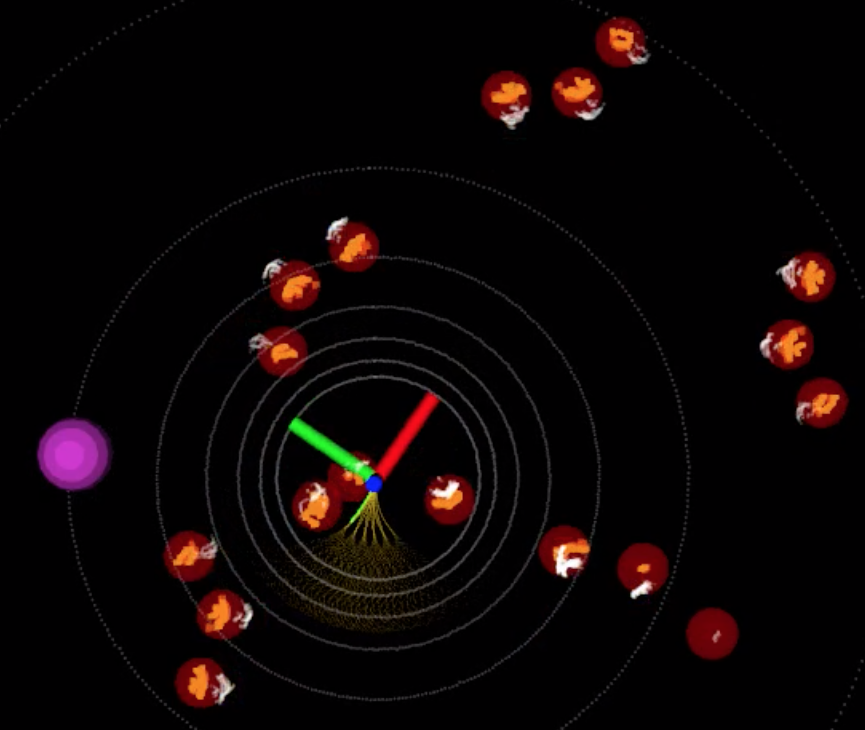}}\;\raisebox{-0.5\height}{\includegraphics[height=3cm,width=.2\linewidth]{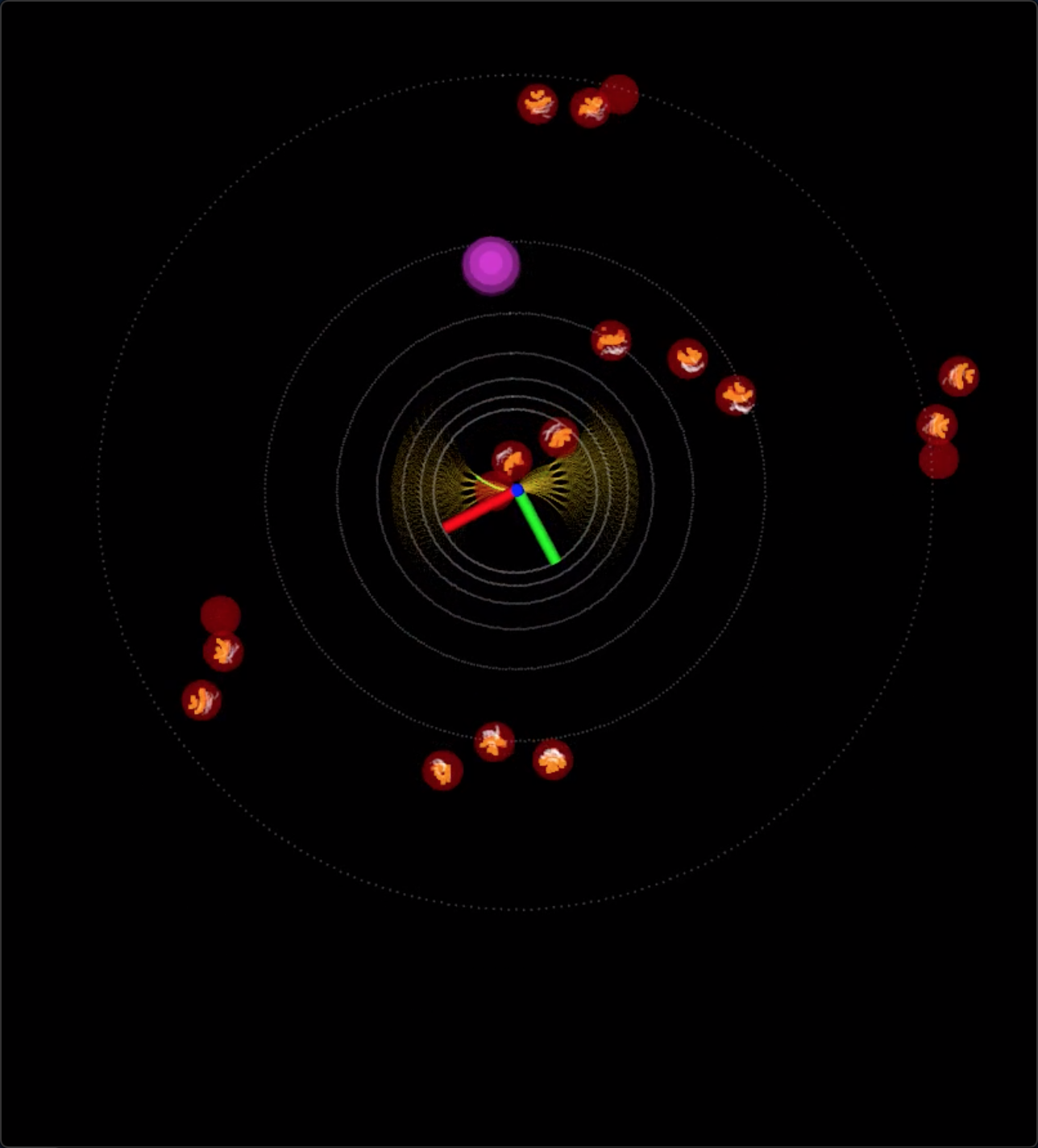}}\;\raisebox{-0.5\height}{\includegraphics[height=3cm,width=.2\linewidth]{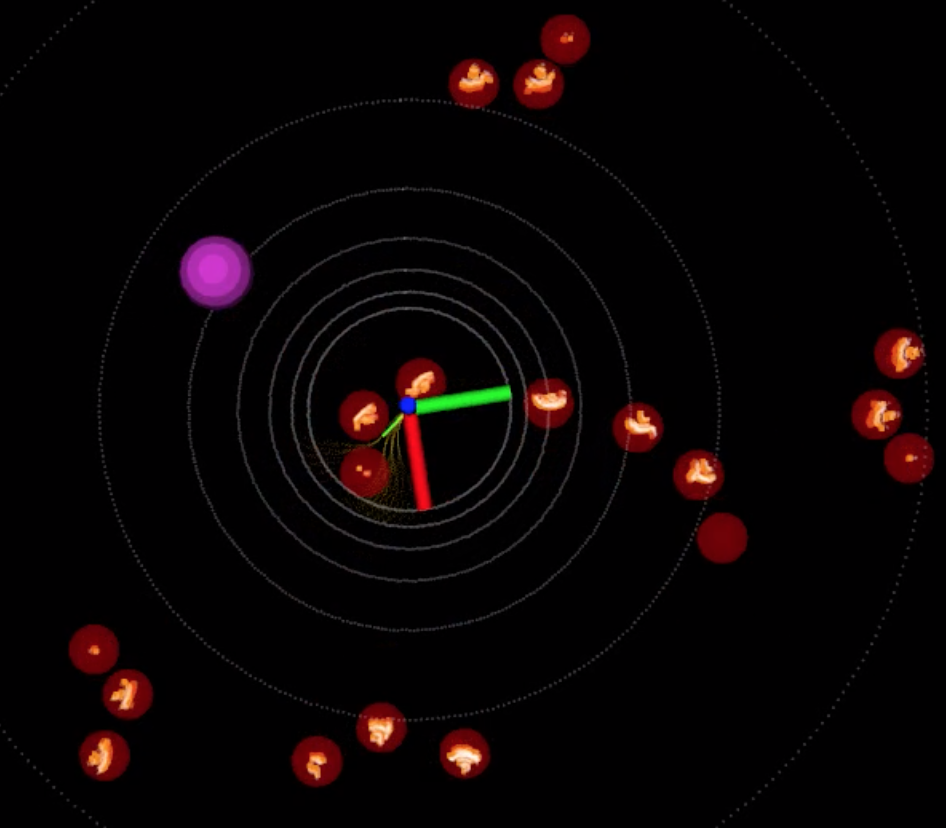}}
    \\\vspace{.3\baselineskip}
    {\rotatebox[origin=c]{90}{\parbox{2.4cm}{{\footnotesize With Social Model}}}}\;\raisebox{-0.5\height}{\includegraphics[height=3cm,width=.2\linewidth]{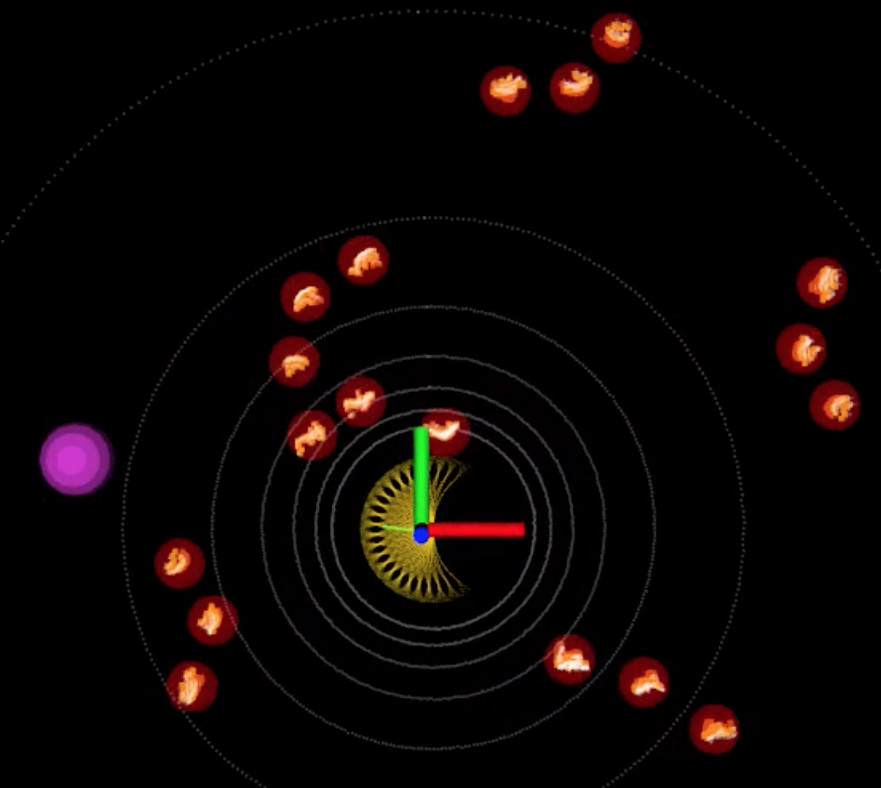}}\;\raisebox{-0.5\height}{\includegraphics[height=3cm,width=.2\linewidth]{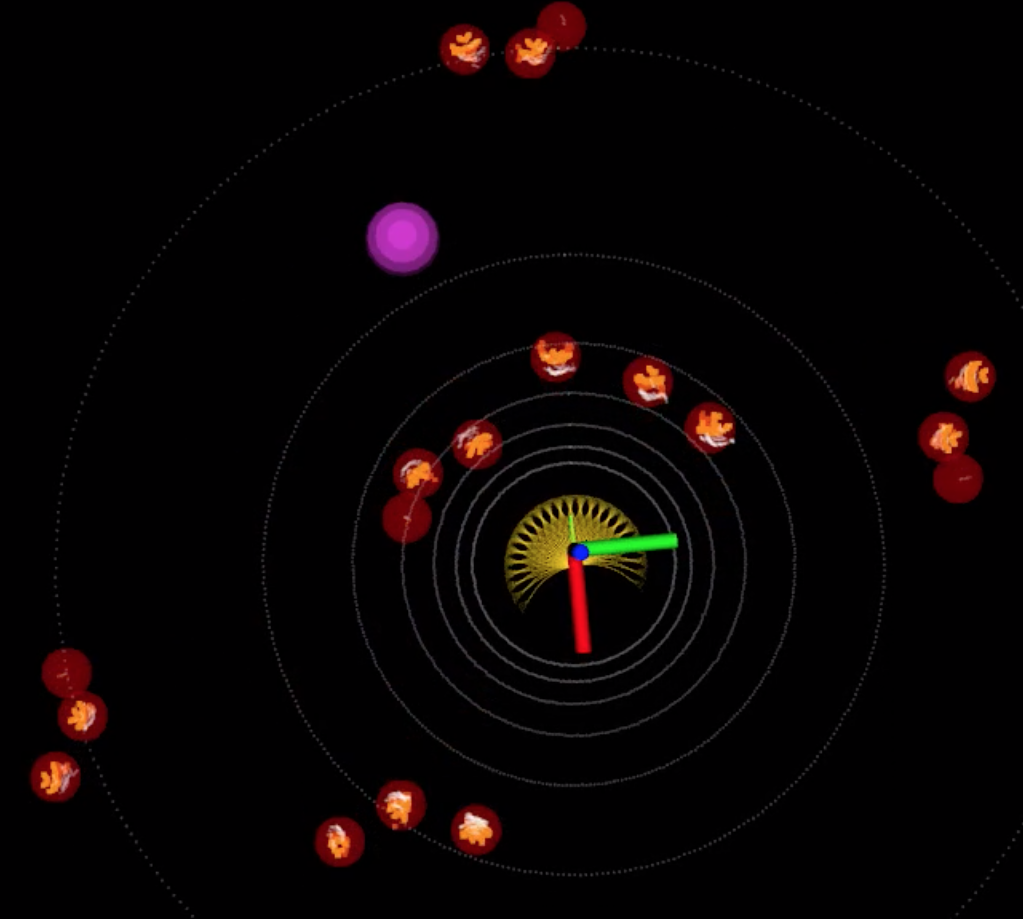}}\;\raisebox{-0.5\height}{\includegraphics[height=3cm,width=.2\linewidth]{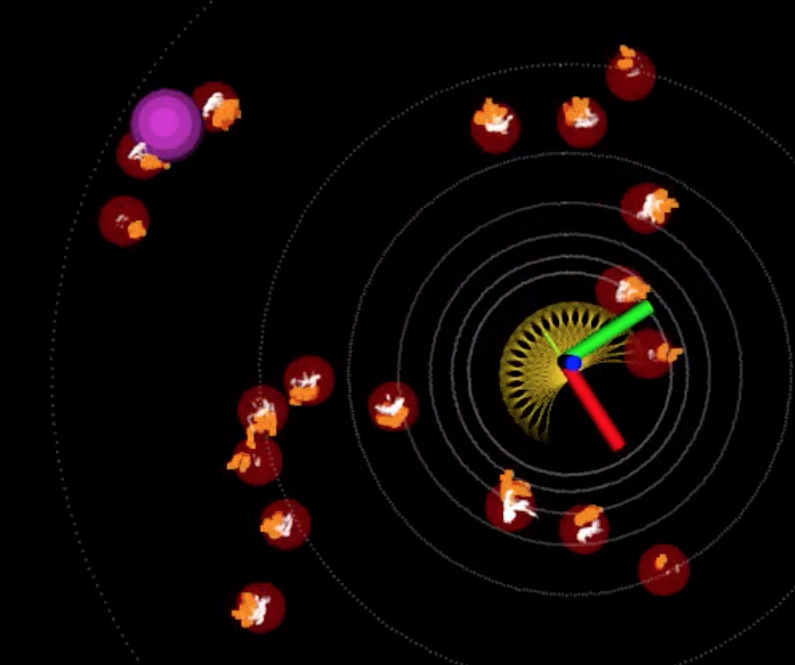}}
    \\\vspace{.3\baselineskip}
     {\rotatebox[origin=c]{90}{\parbox{1.4cm}{{\footnotesize Trajectories}}}}\;\raisebox{-0.5\height}{\includegraphics[height=3.5cm,width=.2\linewidth]{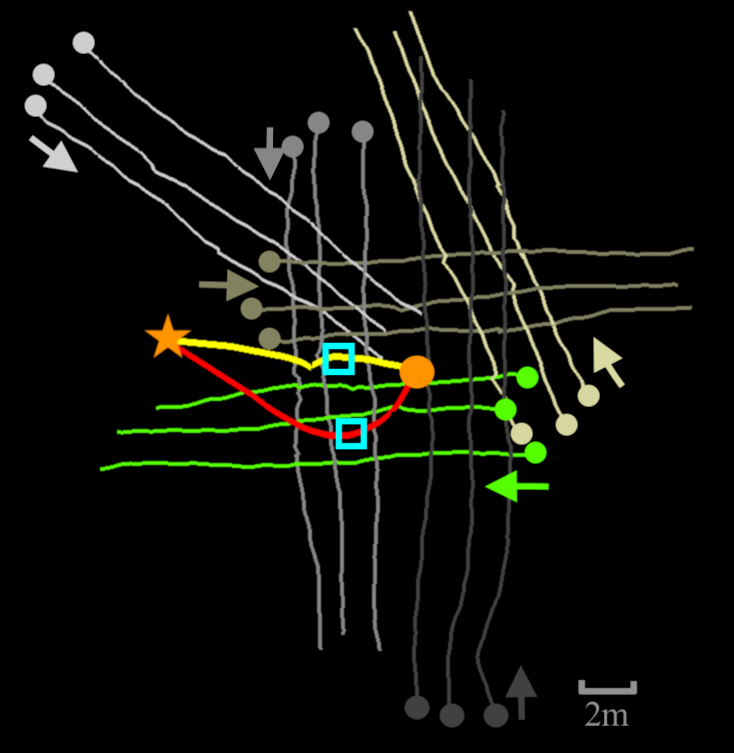}}\;\raisebox{-0.5\height}{\includegraphics[height=3.5cm,width=.2\linewidth]{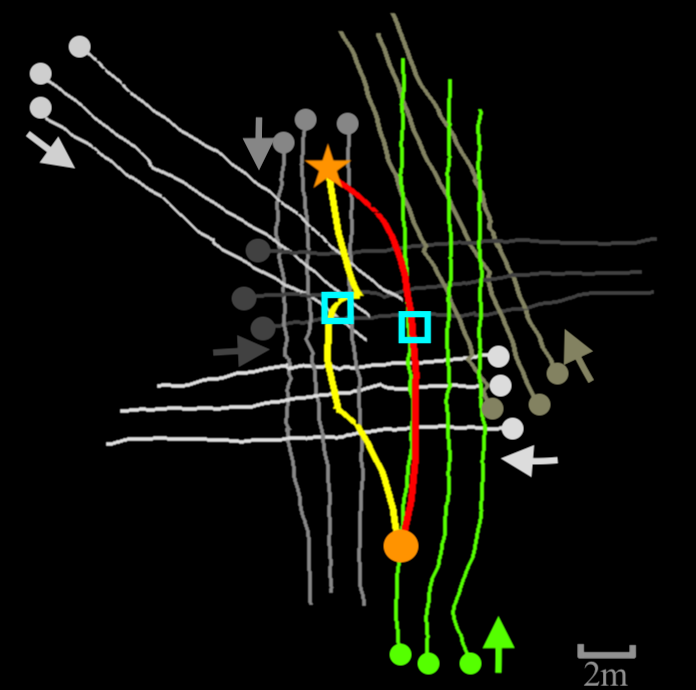}}\;\raisebox{-0.5\height}{\includegraphics[height=3.5cm,width=.2\linewidth]{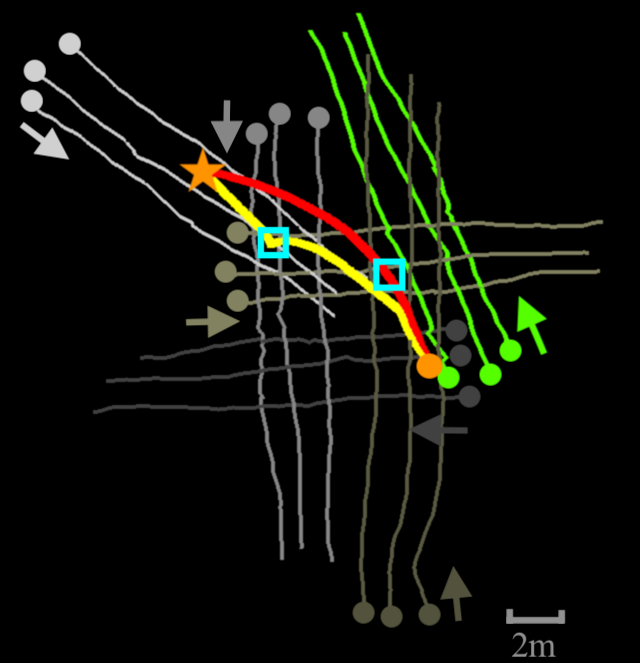}}
    \\\vspace{.2\baselineskip}
    \hspace{0.2cm}{\footnotesize (1)}\hspace{3.3cm}{\footnotesize (2)}\hspace{3.3cm}{\footnotesize (3)}
    \vspace{.2cm}
    \caption{Simulation results in a $\SI{10}{\meter}$ $\times$ $\SI{20}{\meter}$ area. The tests involve 18 people walking in 6 groups. Each group moves in a different direction. The three columns present three representative cases. The first and second rows show screenshots of the simulation environment. The coordinate frame indicates the robot. The goal point is marked as the magenta dot. The red dots are the tracked pedestrians using laser scan data. The third row displays the trajectories of the pedestrians (gray and green) and the robot (yellow and red). The dots are the start points and the star is the goal point of the robot. When using Navigation without Social Model, the robot produces the yellow path. When using Navigation with Social Model, the robot follows the group in green color and produces the red path. A blue square is labeled on each robot path where the corresponding screenshot is captured on the first and second rows. Specifically, on the first row, the screenshots show the moments when the robot drives overly close to people due to not using the social model. On the second row, the screenshots are taken while the robot follows a group during the navigation.}
    \vspace{-.2cm}
\label{fig:simulation}
\end{figure*}

\section{Experiments}

\subsection{Social Compliance Evaluation}

We evaluate our method on two publicly available datasets: ETH \cite{eth} and UCY \cite{ucy}. These datasets include rich social interactions in real-world scenarios. We follow the same evaluation methodology as the leave-one-out approach and the error metrics used in the prior work \cite{socialgan}: 
\begin{enumerate}
\item \textit{Average Displacement Error}:
The average L2 distance between predicted way-points and ground-truth trajectories over the predicted time steps.
\item \textit{Final Displacement Error}:
The L2 distance between the predicted way-point and true final position at the last predicted time step.
\end{enumerate}

We compare against a linear regressor that only predicts straight paths, Social-GAN(SGAN) \cite{socialgan}, and Navi-GAN \cite{navigan}. We use the past eight time steps to predict the future eight time steps. As shown in TABLE \ref{Table:metrics}, our method yields considerable accuracy improvements for some of the datasets where rich group interactions are prevalent. In particular, UNIV and ZARA1 have more than 70\% of the pedestrians moving in social groups, and thus our model performs better. Our model performs slightly worse than the state-of-the-art approaches with the ETH and HOTEL datasets due to the lack of social group interactions. Further, our method assumes the existence of a goal point for each person in the dataset. Lacking precise goal point information results in a relative low accuracy. In the next experiments, we will show results with author-collected data where the strength of our method is more obvious.

\subsection{Group Following Evaluation}

We further evaluate the method with a robot vehicle as shown in Fig.~\ref{fig:vehicle}. The robot is equipped with a Velodyne Puck laser scanner for collision avoidance and pedestrian tracking. Our method is evaluated in two configurations -- Navigation with Social Model refers to the full navigation system as shown in Fig.~\ref{fig:system}, and Navigation without Social Model has the Social Navigation Planning Subsystem removed. The State Estimation Subsystem and the Local Planning Subsystem are directly coupled. The robot navigates directly toward the goal and uses the Local Planning Subsystem to avoid collisions locally.

\setcounter{figure}{3}

\begin{figure}[b!]
  \centering
  \vspace{-.2cm}
  \includegraphics[width=0.5\linewidth]{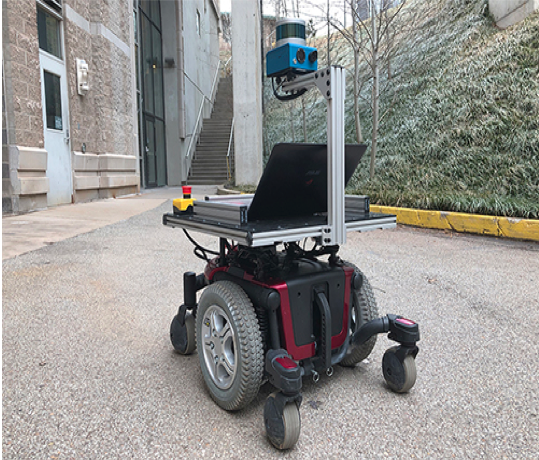}
  \vspace{.2cm}
  \caption{Experiment platform. A wheelchair-based robot carries a sensor pack on the top. The sensor pack consists of a Velodyne Puck laser scanner, a camera, and a low-grade IMU. The scan data is used for collision avoidance and pedestrian tracking. A laptop computer carries out all onboard processing.}
    \label{fig:vehicle}
\end{figure}

\setcounter{figure}{4}

\begin{figure}[h]
\centering
\begin{subfigure}{0.23\textwidth}
  \centering
  \includegraphics[height=2.5cm, width=\linewidth]{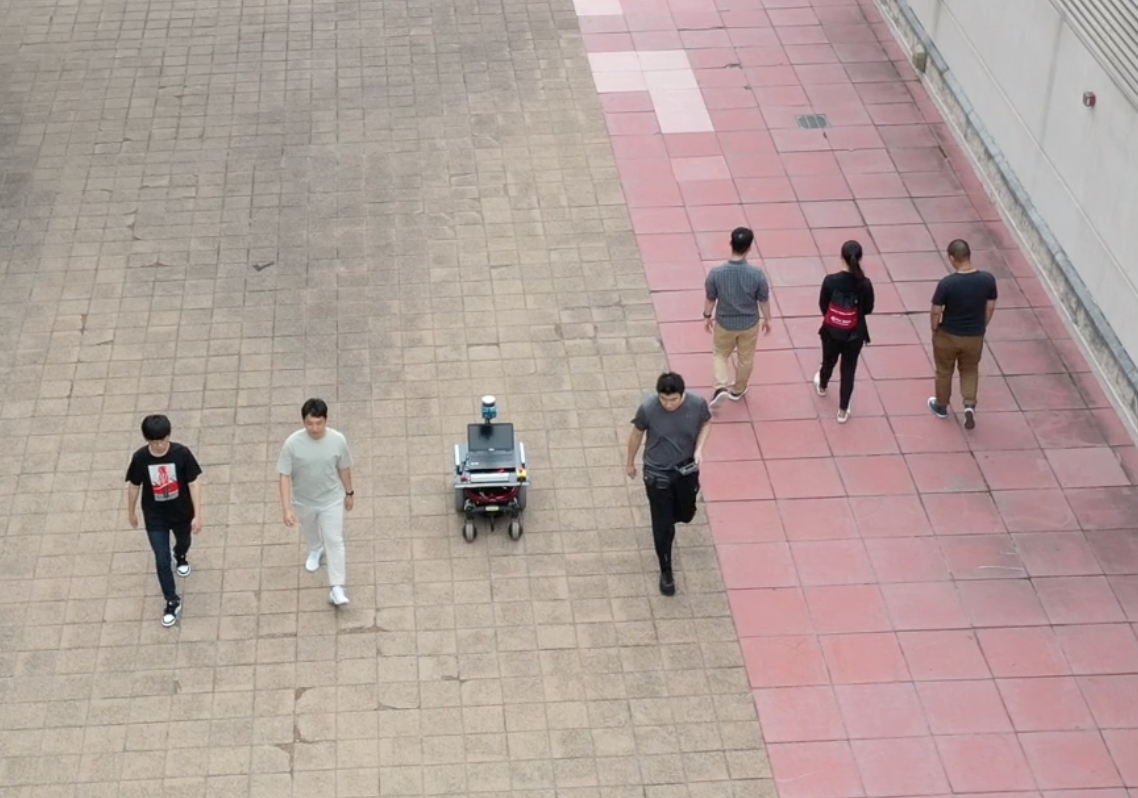}
\end{subfigure}\;
\begin{subfigure}{0.23\textwidth}
  \centering
  \includegraphics[height=2.5cm,, width=\linewidth]{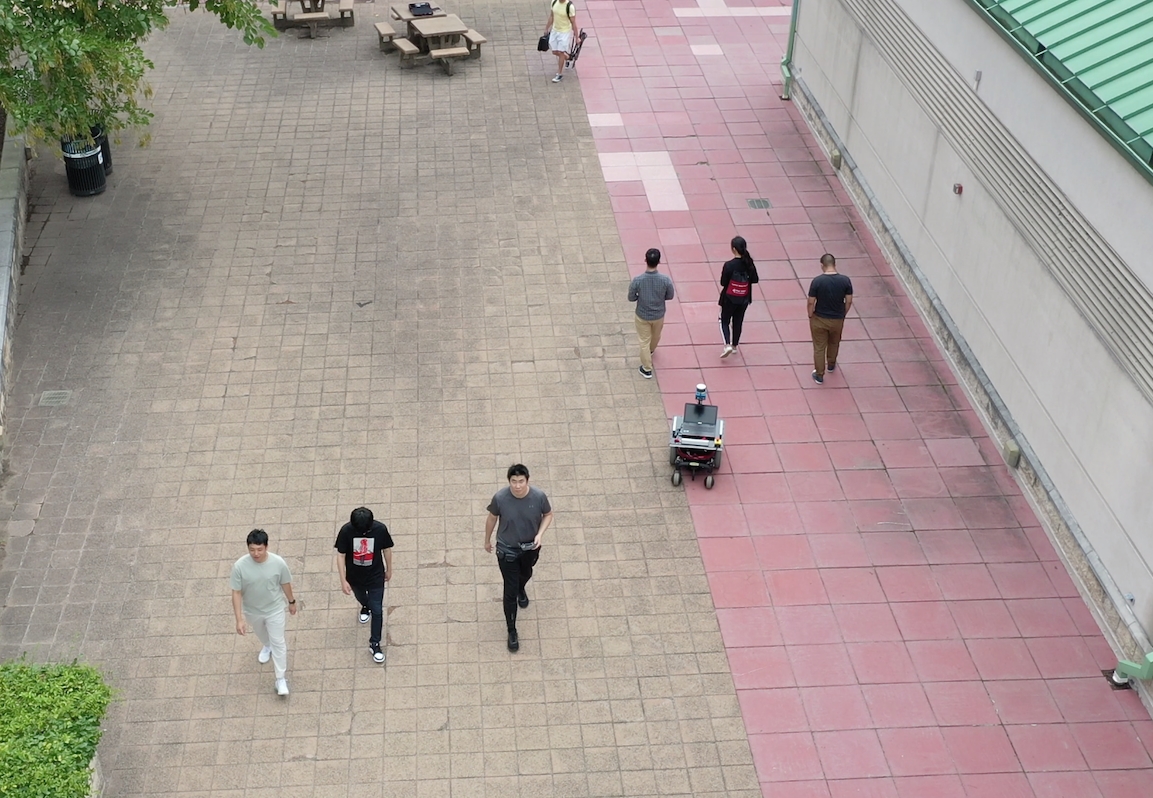}
\end{subfigure} 
\\\vspace{0.3\baselineskip}
\begin{subfigure}[b]{0.23\textwidth}
  \centering
  \includegraphics[height=4cm,width=\linewidth]{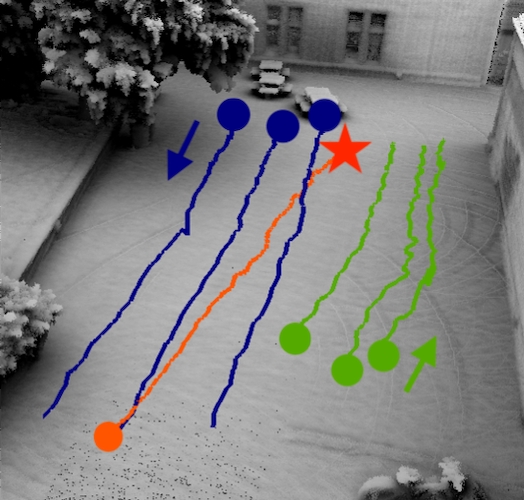}
  \subcaption{\footnotesize Without Social Model}
\end{subfigure}\;
\begin{subfigure}[b]{0.23\textwidth}
  \centering
  \includegraphics[height=4cm,width=\linewidth]{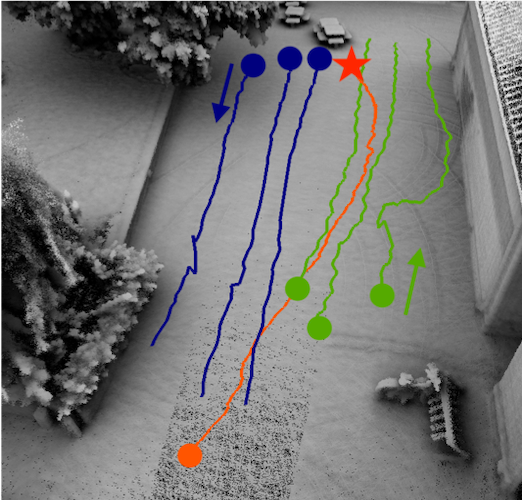}
  \subcaption{\footnotesize With Social Model}
\end{subfigure}
\vspace{.2cm}
\caption{Real-world experiments in a $\SI{10}{\meter}$ $\times$ $\SI{20}{\meter}$ area. The first row shows photos of 6 people walking in 2 groups. One group moves along the robot navigation direction and the other group moves in the opposite direction. The second row shows the corresponding trajectories of the people (blue and green) and the robot (orange). Dots indicate the start points and the star indicates the goal point of the robot. In (a), when using Navigation without Social Model, the robot drives directly toward the goal point and results in cutting through the group on the left that moves against the robot. In (b), when using Navigation with Social Model, the robot follows the group on the right and avoids disturbances to the pedestrians.}
\vspace{-.2cm}
\label{fig:real world}
\end{figure}

We show results in both simulation and real-work experiments with pedestrian data collected by the robot. In simulation, we show scenarios with 18 people walking around the robot in 6 groups. In real-work experiments, we have 6 people walking in 2 groups. One group moves along the robot navigation direction and the other group moves in the opposite direction. The results are shown in Fig.~\ref{fig:simulation} and Fig.~\ref{fig:real world}. In each scenario, the robot selects a group to follow with the full navigation system (Navigation with Social Model). If using Navigation without Social Model, the robot drives directly toward the goal and results in interactions with groups moving in other directions.

Finally, we conduct an Amazon Mechanical Turk (AMT) study to further understand the safety and naturalness of the robot navigation. A total of 466 participants evaluate the simulation and real-world results. As shown in Table~\ref{table:AMT study}, $>$ 90\% of the participants consider Navigation without Social Model to be unsafe (with collisions) while the ratio reduces to $<$ 40\% using Navigation with Social Model. With the real-world results, 95\% of the participants report that the robot forces other pedestrians to change their paths if using Navigation without Social Model. When using Navigation with Social Model, the ratio reduces to 4\%. The survey result validates that our method helps reduce disturbances to other pedestrians as well as improves safety of the navigation. A video of these results can be seen at {\footnotesize\url{www.youtube.com/watch?v=I_SkA9rmxYE}}.

\begin{table}[b]
\centering
\vspace{-.2cm}
\resizebox{0.95\linewidth}{!}{
\begin{tabular}{c|c|c|c}
Metric & Scene & Without Social Model & With Social Model\\
\hline \hline
\textbf{\multirow{4}{*}{Collision (Safety)}} & (1) & 97\%  & 42\% \\
 & (2) & 92\%  & 6\% \\
 & (3) & 92\%  & 36\% \\
\hline
 & \textbf{AVG} &93\%  & 28\% \\
 \hline \hline
\textbf{Path Change (Naturalness)} & Real world & 95\% & 4\% \\
\end{tabular}}
\vspace{.2cm}
\caption{Results of survey study. A total of 466 participants evaluate the simulation results in Fig.~\ref{fig:simulation} and the real-world results in Fig.~\ref{fig:real world}. We can see that $>$ 90\% of the participants consider the Navigation without Social Model to have collisions. For Navigation with Social Model, the ratio reduces to $<$ 40\%. Further, 95\% of the participants report that the robot forces other pedestrians to change their paths if using Navigation without Social Model. When using Navigation with Social Model, the ratio reduces to 4\%. The ratios reduce by 3 times in terms of collision and 20 times in terms of path change which validate that our method helps reduce disturbances to other pedestrians as well as improves safety.}
\label{table:AMT study}
\end{table}

\section{Conclusion}

The paper proposes an autonomous navigation system capable of operating in dense crowds. In this system, a Social Navigation Planning Subsystem incorporating a deep neural network generates socially compliant behaviors. This involves a group pooling mechanism by inferring social relationships to encourage the autonomous navigation to join the flow of a social group sharing the same moving direction. We show the effectiveness of our method through quantitative and empirical studies in both simulations and real-world experiments. The result is that by joining the crowd flow, the robot has fewer collisions with people crossing sideways or walking toward the robot. Joining the flow also creates fewer disturbances to the pedestrians. As a result, the robot navigates in a safe and natural manner. Since this paper focuses on human-robot interactions at a group's level, extension of the work in the future can model interactions between groups and scattered individuals.

\section*{ACKNOWLEDGMENT}
Special thanks are given to C.-E. Tsai, Y. Song, D. Zhao for facilitating experiments.

\bibliographystyle{./IEEEtran}
\bibliography{./IEEEabrv,./IEEEexample}

\end{document}